\documentclass[letterpaper, 10 pt, conference]{ieeeconf}  

\IEEEoverridecommandlockouts                              
\usepackage[utf8]{inputenc}
\usepackage[T1]{fontenc}

\usepackage[style=ieee, sorting=none, citestyle=numeric-comp]{biblatex}
\usepackage[%
colorlinks=True,
citecolor=blue,
pdfborder={0 0 0},
linkcolor=red
]{hyperref}

\usepackage{algorithm} 
\usepackage{algpseudocode}

\usepackage{algorithm}
\usepackage{algorithmicx}
\usepackage{algpseudocode}
\usepackage{amsmath}
\usepackage{amssymb}

\usepackage[utf8]{inputenc}
\usepackage[T1]{fontenc}

\usepackage{graphicx}
\graphicspath{ {./Figures/} }

\usepackage{graphics} 
\usepackage{epsfig} 
\usepackage{mathptmx} 
\usepackage{times} 
\usepackage{amsmath} 

\usepackage{amssymb}  
\usepackage{bm}

\title{\LARGE \bf
Belief-Adaptive Online Autonomy for Quadrotor UAV Navigation under GNSS Degradation in Urban Environments
}

\author{Deepak Kumar Panda$^{1}$ and Weisi Guo$^{1}$
\thanks{*The work is supported by EPSRC
	CHEDDAR: Communications Hub for Empowering Distributed clouD computing Applications and Research (EP/X040518/1) (EP/Y037421/1) and Royal Academy of Engineering UKIC Fellowship}
\thanks{$^{1}$The authors are with Centre for Connected and Assured Autonomy, Cranfield University, Bedford MK43 0AL
        {\tt\small Deepak.Panda, Weisi.Guo @cranfield.ac.uk}%
}}

\begin{document}

\maketitle
\thispagestyle{empty}
\pagestyle{empty}

\begin{abstract}
Reliable online autonomy is critical for quadrotor operation in urban airspaces, where global navigation satellite systems (GNSS) measurements suffer from multipath, blockage, and latency issues, introducing non-stationary, temporally correlated errors that degrade conventional GNSS–IMU fusion. This paper presents a belief-adaptive online autonomy framework that augments an extended Kalman filter (EKF) with explicit GNSS trust modelling, second-order online belief adaptation, and latency-aware out-of-sequence measurement handling. GNSS trust is represented as a latent belief state that modulates measurement weighting and multipath bias uncertainty, and is updated online using EKF consistency signals. Unlike reactive covariance tuning, the proposed approach enables proactive and stable sensor trust adaptation without prior environmental knowledge or offline training. Evaluation in simulated urban air mobility scenarios with correlated multipath, stochastic latency, and obstacle constraints demonstrates improved belief convergence, smoother trajectories, and reduced estimation and tracking errors compared to naive, adaptive, and first-order baselines. The framework preserves classical GNSS-IMU fusion structure and can be integrated directly into existing flight-control pipelines, supporting robust online autonomy in GNSS-degraded environments.
\end{abstract}

\section{INTRODUCTION}
Urban air mobility (UAM) places stringent requirements on the navigation reliability of quadrotor unmanned aerial vehicles (UAVs) operating in dense urban airspaces \cite{isik2024machine}. Quadrotors, navigating in an integrated airspace, must maintain accurate and resilient positioning while flying close to urban infrastructures \cite{fan2022quadrotor}. Global navigation satellite systems (GNSS) remain the primary source of position, navigation, and timing (PNT) information for quadrotor flight control \cite{chi2021enabling, negru2025resilient}, offering high accuracy under open-sky conditions. However, in urban environments GNSS performance degrades due to signal blockage, multipath reflections \cite{geragersian2022multipath}, interference, and latency \cite{gamagedara2021quadrotor}. These effects introduce non-line-of-sight propagation and non-stationary errors \cite{negru2023realism} that violate the assumptions of conventional navigation filters and pose a major challenge for reliable UAM autonomy.

To improve robustness, GNSS is commonly fused with inertial measurement units (IMUs) in loosely or tightly coupled architectures \cite{chi2021enabling, qin2014performance, sun2021new}. IMUs provide high-rate motion information that enables quadrotors to bridge short GNSS outages and maintain smooth state estimates. However, IMU-only dead reckoning accumulates bias and scale-factor errors, making periodic GNSS correction essential for long-term accuracy. As a result, GNSS–IMU fusion remains the baseline navigation architecture for quadrotor UAM platforms.

The performance of GNSS–IMU fusion is dependent on the underlying estimation algorithms like Extended Kalman filters (EKF) \cite{iyer2024enhancing} and unscented Kalman filters \cite{yang2018correlational}, which perform approximate Bayesian inference under assumed process and measurement noise distribution. In urban environments, GNSS errors vary abruptly due to multipath, blockage, and latency \cite{gong2020graph}, yet are often modeled using fixed or slowly tuned covariances. This assumption of static sensor reliability can lead to position estimation errors that may compromise mission safety under unpredictable GNSS conditions.

Adaptive filtering approaches have been proposed to mitigate non-stationary GNSS errors, including innovation and residual-based covariance tuning \cite{li2013adaptive}, redundant noise estimation \cite{yin2023robust}, variational Bayesian filtering \cite{ma2023variational}, and robust heavy-tailed formulations \cite{yu2024robust}. Despite their differences, these methods share a common structural assumption that sensor reliability is inferred reactively from innovation or residual statistics at the current time step, while not capturing the measurement degradation explicitly. However, in urban GNSS environments these assumptions break down, as multipath effects, signal blockages and latencies generate abrupt, geometry-dependent, and temporally correlated errors. As a result, innovation-based covariance adaptation often responds only after large errors have already corrupted the state estimate, leading to delayed recovery and inconsistent estimator confidence. Redundant noise estimation and variational Bayesian methods \cite{yin2023robust,ma2023variational} assume local stationarity, limiting adaptation to rapidly varying GNSS degradation. Heavy-tailed filters \cite{yu2024robust} mitigate isolated outliers but struggle with persistent multipath bias and delayed measurements that break temporal alignment assumptions. More importantly, these methods lack an explicit mechanism to represent and adapt measurement trust over time, preventing sustained modulation of filter behavior. Machine-learning-based navigation methods have also been explored to model nonlinear and context-dependent GNSS errors \cite{tabassum2023integrating, geragersian2022ins}. While effective in specific settings, such approaches introduce training dependence, additional computational overhead, and sensitivity to out-of-distribution conditions, posing challenges to quadrotor autonomy in UAM. Collectively, these limitations motivate lightweight online adaptation mechanisms that preserve the structure and interpretability of classical estimation pipelines, while enabling proactive adjustment of sensor trust under non-stationary urban GNSS degradation.

In this paper, a second-order online belief-adaptive EKF is proposed for quadrotor navigation under GNSS degradation. The framework represents GNSS reliability as an explicit latent belief state that governs both measurement weighting and multipath-bias dynamics, and updates this belief online using a curvature-aware adaptation rule driven by filter consistency signals. Unlike conventional adaptive filters that adjust sensor confidence only after large residuals have occurred, the proposed approach proactively modulates GNSS trust, yielding improved performance under non-stationary conditions. In contrast to ML-based sensor fusion techniques, the proposed online learning mechanism operates without offline training on environment-specific data \cite{tabassum2023integrating, panda2024action}, enabling data-efficient, and robust adaptation suitable for UAM operations. In addition, a latency-aware out-of-sequence update mechanism, similar to the approach in \cite{gamagedara2021quadrotor}, is integrated to retroactively correct delayed measurements and re-propagate state estimates, ensuring temporally consistent GNSS–IMU fusion in the presence of measurement latency. This paper does not propose a new planner, controller or GNSS degradation model; they serve as a controlled experimental environment to isolate and evaluate the effect of second-order belief adaptation within the estimation layer. This paper contributes only to the estimator adaptation layer; all other modules remain fixed across comparisons. The paper contributions can be summarized as follows:
\begin{itemize}
	\item \textbf{Curvature-Aware Second-order Online Belief Adaptation for GNSS Trust Modulation:} 
	We introduce a dual-timescale second-order online Newton update that explicitly models GNSS reliability as a latent belief state governing both measurement covariance and bias process dynamics. Unlike innovation-based covariance tuning, the proposed mechanism adapts trust proactively using curvature information, yielding improved dynamic navigation under non-stationary degradation.
	\item \textbf{Evaluation in Urban Air Mobility Environments:} The framework is evaluated in a sample urban air mobility simulation framework with geometry-dependent multipath bias and correlated latency, while comparing against vanilla EKF, non-belief aware adaptive EKF with ablation study comparison with first-order methods.
\end{itemize}
\begin{figure*}[thpb]
	\centering
	\includegraphics[scale=0.45]{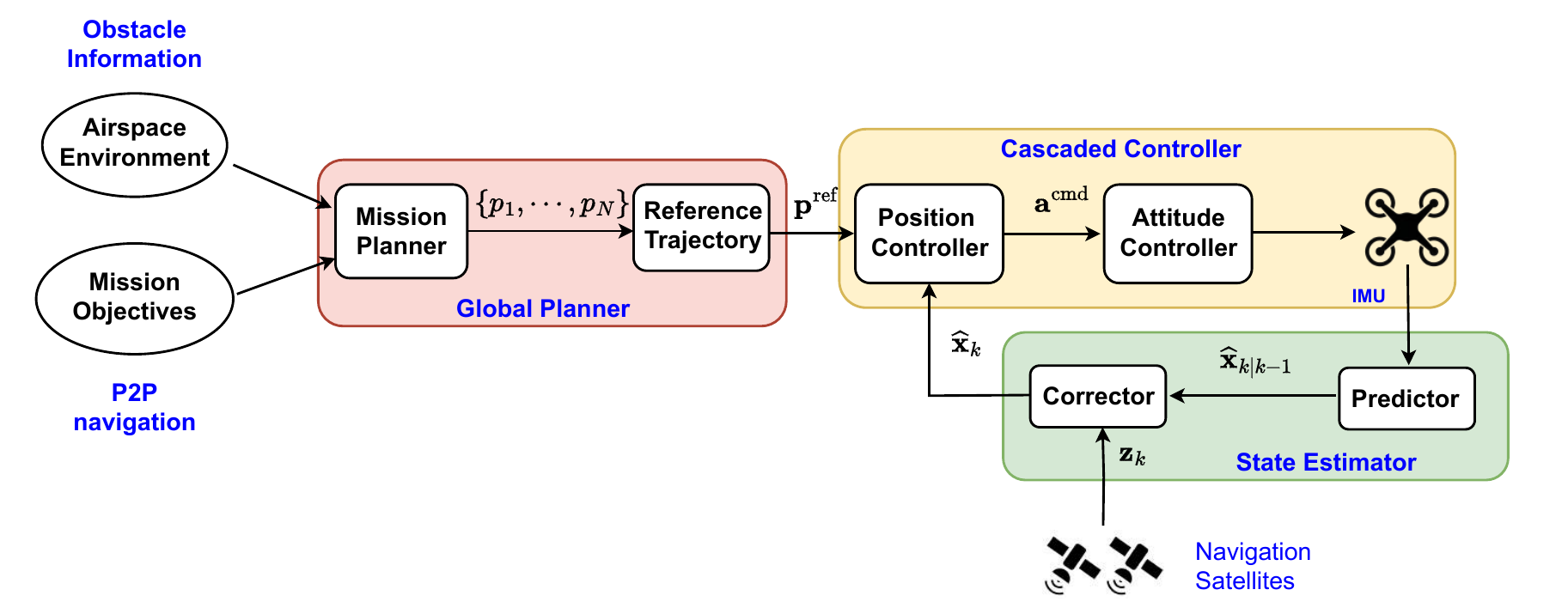}
	\caption{Overview of the closed-loop mission planning, estimation, and control architecture for autonomous UAV navigation.}
	\label{figure1_schematic}
\end{figure*}
\section{Quadrotor Navigation System}
Figure \ref{figure1_schematic} illustrates an integrated closed-loop architecture for autonomous point-to-point UAV navigation in urban environment. The quadrotor follows standard nonlinear rigid-body dynamics \cite{beard2012small} with a cascaded position-attitude controller. The mission objectives and obstacle information in the urban airspace environment are processed by a global mission planner to generate a sequence of waypoints ${p_1, \cdots, p_N}$ and subsequently minimum jerk reference trajectory $\mathbf{p_{\text{ref}}}$. This reference trajectory is tracked by a cascaded position–attitude controller, which computes low-level control commands using the feedback of the estimated states from EKF. In parallel, a predictor–corrector state estimator as, a part of EKF, fuses high-rate IMU data with GNSS measurements $\mathbf{z}_j$ from navigation satellites to estimate the quadrotor position $\hat{\mathbf{x}}_k$, and the resulting estimated position is fed back to the controller to enable accurate and robust trajectory tracking under sensing uncertainty.
\subsection{Global Path Planning in Urban Environment}
The operational environment is represented as a 2D binary occupancy grid $\mathcal{G} \subset \mathbb{Z}^2$, where buildings are modeled as axis-aligned rectangular obstacles, consistent with urban GIS abstractions and voxelized city maps. Obstacle clearance is quantified using an Euclidean distance transform (EDT) and incorporated directly into the global planning cost.
\begin{equation}\label{eq:1}
d_{\text{obs}}(c) = \min_{c' \in \mathcal{O}} \lVert c - c' \rVert_2,
\end{equation}
where $\mathcal{O}$ denotes the set of occupied cells due to obstacles. Modified A* is used as a global planner on the grid representing urban environment, while considering the curvature-aware costs that affect downstream quadrotor trajectory tracking. Total evaluation function for the global planner is defined as, 
\begin{equation}\label{eq:2} 
	f \left ( n \right) = g \left ( n \right) + h \left ( n \right),
\end{equation}
where $g \left ( n \right)$ is the accumulated path cost, and $h \left ( n \right)$ is the admissible heuristic. The path cost at node $n_k$, $g \left (n_k \right)$, depends on the previous cost $g \left (n_{k-1} \right)$ and the incremental cost $g_{\text{inc}} \left ( \cdot \right )$.The criteria for expanding from node $n_{k-1}$ to $n_k$ considers the:
\begin{itemize}
	\item Traversal cost enforcing path length minimization $g^{\text{tra}}_{\text{inc}} \left ( \cdot \right )$.  
	\item Regularization cost which discourages aggressive turns that are dynamically infeasible for quadrotors $g^{\text{turn}}_{\text{inc}} \left ( \cdot \right )$.
	\item Clearance cost representing the distance from the nearest obstacle $g^{\text{clr}}_{\text{inc}} \left ( \cdot \right )$.
\end{itemize}
Hence, the total accumulated path cost for the quadrotor tracking is given by:
\begin{equation} \label{eq:3}
	f \left ( n \right) = w^{\text{tra}} \cdot g^{\text{tra}}_{\text{inc}} \left ( \cdot \right ) + 
						  w^{\text{turn}} \cdot g^{\text{turn}}_{\text{inc}} \left ( \cdot \right ) + 
						  w^{\text{clr}} \cdot g^{\text{clr}}_{\text{inc}} \left ( \cdot \right ) + 
						  h \left ( n \right).
\end{equation}
The heuristic $h \left (n \right)$ is chosen as the Euclidean distance to the goal, given by, $h(n) = \lVert p_n - p_{\text{goal}} \rVert_2.$ After solving the given global planning algorithm resulting plan is a sequence of waypoints given by
\begin{equation}\label{eq:4}
\mathcal{P}	= \left \{ p^{\text{wp}}_0, p^{\text{wp}}_1, \cdots, p^{\text{wp}}_N \right \}.
\end{equation}
Based on the waypoints obtained in (\ref{eq:4}), the next task is for the quadrotor control system to track the straight line, based on the reference trajectory generated as given in the next subsection.
\subsection{Reference Trajectory Generation}
Based on the discrete sequence of the waypoints given in (\ref{eq:4}), a time-parametrized minimum-jerk trajectory is generated between each consecutive waypoint pair $\left ( p^{\text{wp}}_i,  p^{\text{wp}}_{i+1} \right ).$ For each segment, motion is parameterized over a fixed horizon $T_i$ using a normalized time variable $\tilde{s} = t / T_i, \quad \tilde{s} \in \left [ 0,1 \right ]$. The position trajectory is defined as,
\begin{equation}\label{eq:5}
	\mathbf{p} \left ( \tilde{s} \right ) = \left (1 - \alpha \left (\tilde{s} \right ) \right )\, \mathbf{p}_i + \alpha \left ( \tilde{s} \right )\, \mathbf{p}_{i+1},
\end{equation}
where $\alpha \left ( \tilde{s} \right ) = 10\tilde{s}^3 - 15\tilde{s}^4 + 6\tilde{s}^5$ is classical minimum-jerk polynomial. Similarly, reference velocity and acceleration can be obtained by differentiating the position trajectory given in (\ref{eq:5}) accordingly. The reference trajectories ensure smooth continuity with bounded acceleration and no jerk discontinuities. 
\subsection{Quadrotor Dynamics}
We consider a rigid body quadrotor operating at low altitude in an urban environment. The full state vector is defined as per \cite{beard2012small},  
\begin{equation}\label{eq:6}
	\mathbf{s} = \left [ x,y,z,v_x, v_y, v_z, \varphi, \vartheta, \tilde{\psi}, p, q, r \right ].
\end{equation}
Here, $\left (x,y,z \right)$ denote inertial-frame position, $\left (v_x, v_y, v_z \right )$ represents linear velocity, $\left (\varphi, \vartheta, \tilde{\psi} \right)$ represent roll, pitch and yaw angles and $\left (p,q,r \right )$ represents body-frame angular rates.
\begin{equation}\label{eq:7}
	\dot{\mathbf{p}} = \mathbf{v}, \qquad
	\dot{\mathbf{v}} = \frac{1}{m}\,\mathbf{\varrho}(\varphi,\vartheta, \tilde{\psi})
	\begin{bmatrix}
		0\\
		0\\
		T
	\end{bmatrix}
	-
	\begin{bmatrix}
		0\\
		0\\
		g_g
	\end{bmatrix},
\end{equation}
where $m$ represent the quadrotor mass, $T$ is the total thrust, $g_g$ being the gravitational acceleration and $\mathbf{\varrho} \left( \cdot \right)$ is the rotational matrix from the body to the inertial frame. The rotation matrix $\mathbf{\varrho}$ in (\ref{eq:6}) is defined as,
\[
\mathbf{\varrho} =
\begin{bmatrix}
	c_\psi c_\theta
	& c_\psi s_\theta s_\phi - s_\psi c_\phi
	& c_\psi s_\theta c_\phi + s_\psi s_\phi \\[4pt]
	s_\psi c_\theta
	& s_\psi s_\theta s_\phi + c_\psi c_\phi
	& s_\psi s_\theta c_\phi - c_\psi s_\phi \\[4pt]
	- s_\theta
	& c_\theta s_\phi
	& c_\theta c_\phi
\end{bmatrix}.
\]
Here, $c \left ( \cdot \right)$ and $s \left ( \cdot \right)$ represent cosine and sine functions.
\subsubsection{Rotational Dynamics}
The rotational dynamics follow rigid-body equations shown as follows
\begin{equation}\label{eq:8}
	\dot{\boldsymbol{\omega}}
	= \mathbf{I}^{-1}\!\left(
	\boldsymbol{\tau}
	- \boldsymbol{\omega} \times (\mathbf{I}\boldsymbol{\omega})
	\right),
\end{equation}
where $\boldsymbol{\omega} = \left [ p, q,r \right ]^T$, $\boldsymbol{\tau} = \left [ \tau_{\varphi}, \tau_{\vartheta}, \tau_{\tilde{\psi}} \right ]^T$ and $\mathbf{I}$ is the inertia matrix. Euler angles are propagated via first-order integration of body rates. The full nonlinear quadrotor model is discretized at IMU rate of 200 Hz for simulation and estimation propagation.
\subsubsection{Cascaded Control Systems}
Trajectory tracking is achieved using a cascaded outer-loop position controller and inner-loop attitude controller, operating on the estimated state provided by the onboard state estimator \cite{beard2012small}. The controller-estimator structure ensures system stability in case of latencies and data corruption in safety-critical environment similar to \cite{panda2024observer}. Given the reference trajectory $(p_{\text{ref}},\, \dot{p}_{\text{ref}},\, \ddot{p}_{\text{ref}})$, the commanded acceleration is computed as,  
\begin{equation}\label{eq:9}
a_{\text{cmd}}
= \ddot{\mathbf{p}}_{\text{ref}}
+ K_p \bigl(\mathbf{p}_{\text{ref}} - \hat{\mathbf{p}}\bigr)
+ K_d \bigl(\dot{\mathbf{p}}_{\text{ref}} - \hat{\mathbf{v}}\bigr)
+ K_i \int \bigl(\mathbf{p}_{\text{ref}} - \hat{\mathbf{p}}\bigr)\, dt.
\end{equation}
where $\hat{\mathbf{p}}$ and $\hat{\mathbf{v}}$ are the estimated position and velocity from EKF, and $K_p, K_i, K_d$ are proportional, integral and derivative gain to track the reference trajectory. To prevent the integrator windup, the integral term is activated when position reference and value is within a given threshold $\| p_{\mathrm{ref}} - \hat{p} \| < \varepsilon_{\mathrm{int}}$. The commanded acceleration vector is mapped to desired roll and pitch angles under the small-angle approximation to generate attitude references provided as follows:
\begin{equation}\label{eq:10}
	\varphi_{\text{cmd}} = -\frac{a_{\text{cmd}, y}}{g_g}, \quad \vartheta_{\text{cmd}} = -\frac{a_{\text{cmd}, x}}{g_g}.
\end{equation}
The commanded thrust is provided as $T_{\text{cmd}} = m \left ( g_g + a_{\text{cmd}, z} \right)$. The desired yaw angle is held constant or aligned with the heading to minimize the lateral drag. The inner-loop stabilization is achieved via PD control with angular-rate damping. The control torques are provided as
\begin{equation}\label{eq:11}
	\begin{gathered}
		\tau_{\varphi} = k_{\varphi}\left(\varphi_{\text{cmd}} - \varphi\right) - d_{\varphi} \, p ,\\
		\tau_{\vartheta} = k_{\vartheta}\left(\vartheta_{\text{cmd}} - \vartheta\right) - d_{\vartheta} \, q. 
	\end{gathered}
\end{equation}
The cascaded structure ensures time-scale separation, allowing the inner loop to stabilize attitude dynamics significantly faster than the outer-loop translational controller.
\section{Online Autonomy Framework}
\begin{figure}[thpb]
	\centering
	\includegraphics[scale=0.50]{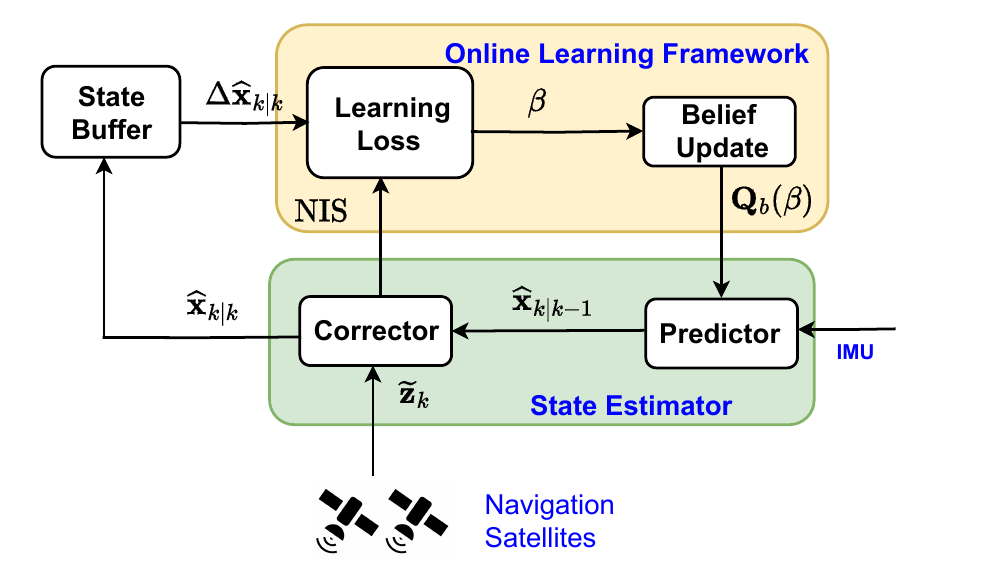}
	\caption{Schematic of the belief-adaptive online learning framework for UAV navigation.}
	\label{figure2_learning_framework}
\end{figure}

As shown in Fig. \ref{figure2_learning_framework}, the predictor–corrector EKF is augmented with a belief-driven adaptation layer that models GNSS reliability in real time. Innovation consistency statistics are used to update a scalar belief variable $\beta$, which modulates measurement covariance to proactively adjust sensor trust. The update does not increase the EKF state dimension and adds negligible computational overhead, preserving real-time feasibility. A state buffer supports temporally consistent and out-of-sequence measurement handling, ensuring stable closed-loop estimation under GNSS degradation. 
\subsection{Extended Kalman Filter}
\subsubsection{Process Model}
We consider the EKF maintaining an augmented state-vector $\mathbf{x}_k \triangleq \begin{bmatrix}
	\mathbf{p}_k & \mathbf{v}_k & \mathbf{b}_k
\end{bmatrix}^T \in \mathbb{R}^{1\times 9}
$. Here $\mathbf{x}_k, \mathbf{v}_k$ represent the position and velocity estimate, and $\mathbf{b}_k$ represents the estimated bias due to GNSS degradation. The overall process model is given by,
\begin{equation}\label{eq:12}
	\mathbf{x}_{k+1} = \mathbf{F} \mathbf{x}_{k} + \mathbf{B} \mathbf{a_k} + \mathbf{w}_k, 
\end{equation}
with $\mathbf{F} =
\begin{bmatrix}
	\mathbf{I} & \Delta t\, \mathbf{I} & \mathbf{0} \\
	\mathbf{0} & \mathbf{I}           & \mathbf{0} \\
	\mathbf{0} & \mathbf{0}           & \alpha \mathbf{I}
\end{bmatrix},
\;
\mathbf{B} =
\begin{bmatrix}
	\frac{1}{2}\Delta t^{2}\, \mathbf{I} \\
	\Delta t\, \mathbf{I} \\
	\mathbf{0}
\end{bmatrix}.$ The process noise $\mathbf{w}_k$ is a zero-mean random disturbance capturing unmodelled dynamics, and modelling uncertainty in the propagation step due to potential GNSS degradation here. We consider $\mathbb{E} \left [ \mathbf{w}_k \right ] = 0, \quad \mathbb{E} \left [ \mathbf{w}^T_k \mathbf{w}_k \right ] = \mathbf{Q}_k. $. We can decompose that as 
\begin{equation}\label{eq:13}
\mathbf{Q}_k =
\begin{bmatrix}
	\mathbf{Q}_{pv} & \mathbf{0} \\
	\mathbf{0} & \mathbf{Q}_b(\beta_k)
\end{bmatrix} 
\quad
\mathbf{Q}_{pv}
= \sigma_a^2
\begin{bmatrix}
	\dfrac{\Delta t^3}{3}\,\mathbf{I} & \dfrac{\Delta t^2}{2}\,\mathbf{I} \\
	\dfrac{\Delta t^2}{2}\,\mathbf{I} & \Delta t\,\mathbf{I}
\end{bmatrix}.
\end{equation}
Here, $\mathbf{Q}_b \left ( \beta_k \right ) = \left ( 1 - \alpha^2 \right ) \sigma^2_b \left ( \beta \right ) \mathbf{I}$, with $\sigma_b \left ( \beta_k \right) = \left ( 1 - \beta_k  + \epsilon_k \right )$. Here $\beta_k$ is the belief parameter which has to be adapted thus enabling the estimator to respond quickly to GNSS degradation via covariance matrix $\mathbf{Q}_k$
\subsubsection{Measurement Model}
GNSS provides position measurement corrupted by multipath bias and latency given by $\mathbf{\hat{z}}_k = \mathbf{p}_k + \mathbf{b}_k + \mathbf{\nu}_k$.  If $\hat{\mathbf{x}_{k \mid k-1}}$ represents the predicted state from the process model $\mathbf{x}_k$, we can compute the normalized innovation square (NIS) from the measurement residuals as,
\begin{equation}\label{eq:14}
\begin{gathered}
\mathbf{y}_k = \mathbf{z}_k - \mathbf{H} \hat{\mathbf{x}_{k \mid k-1}}, \\
\mathbf{S}_k = \mathbf{H} \mathbf{P}_{k \mid k-1} \mathbf{H}^T + \mathbf{R}_k \\
\text{NIS}_k = \mathbf{y}^T_k \mathbf{S}^{-1}_k \mathbf{y}_k
\end{gathered}
\end{equation}
\subsection{Online Learning Framework}
Classical adaptive EKFs, adjust measurement covariance reactively from instantaneous innovations, assuming locally informative and weakly correlated residuals, which leads to delayed and unstable trust adaptation under geometry-dependent multipath and latency. To address this, we introduce an explicit scalar belief state $\beta_k $ as a structural augmentation of the EKF, as per (\ref{eq:12}), that modulates GNSS trust via covariance adaptation, updated via a curvature-aware second-order online learning. The proposed dual-timescale curvature accumulation approximates an online Newton step for fast, stable, and bounded trust adaptation parametrized by a sigmoid function as follows,
\begin{equation}\label{eq:15}
\beta_k = \sigma \left ( \theta_k \right ) = \frac{1}{1 + e^{-\theta_k}}.
\end{equation}
In addition to the innovation consistency, abrupt changes in the estimated state may indicate instability caused by corrupted measurements $\mathbf{\tilde{z}}_k$. Hence, a hybrid loss function is considered as, 
\begin{equation}\label{eq:16}
\ell_k = \log\!\left( 1 + \frac{\mathrm{NIS}_k}{\chi^2_{0.95}} \right)
+ \lambda \left\lVert \hat{p}_{k|k} - \hat{p}_{k-1|k-1} \right\rVert_2, 
\end{equation}
where the first term penalizes statistical inconsistency and the second discourages rapid fluctuations in the estimated position. The gradient of the loss with respect to $\theta_k$, using second order Newton methods, is given by:
\begin{equation}\label{eq:17}
	\nabla_{\theta} l_k = - \frac{\delta l}{\delta \beta} \beta_k \left ( 1- \beta_k \right ), 
\end{equation}
where $\frac{\delta l}{\delta \beta}$ is obtained through differentiation of (\ref{eq:16}) with respect to the belief-dependent covariance terms. We maintain two second-moment accumulators of the gradient magnitude:
\begin{equation}\label{eq:18}
\begin{aligned}
H_k^{\text{fast}} &= \alpha_f H_{k-1}^{\text{fast}} + {\nabla_{\theta} l_k}^2, \\
H_k^{\text{slow}} &= \alpha_s H_{k-1}^{\text{slow}} + {\nabla_{\theta} l_k}^2, 
\end{aligned}
\end{equation}
where $0 < \alpha_f \ll \alpha_s < 1$.  Here $H_k^{\text{fast}}$ captures short-term curvature, and $H_k^{\text{slow}}$ captures long-term curvature in the loss-function. The effective curvature for learning is a convex combination given as:
\begin{equation}\label{eq:19}
	H_k = \omega_f H_k^{\text{fast}} + \left ( 1 - \omega_f \right ) H_k^{\text{slow}}, \qquad \omega_f \in \left (0,1 \right ).
\end{equation}
This dual-timescale design allows the learning process to be both reactive and stable, avoiding oscillations that arise from purely first-order methods. The belief parameter is updated using a damped online Newton-type rule, given as,
\begin{equation}\label{eq:20}
	\theta_{k+1} = \theta_k - \eta \frac{\nabla_{\theta} l_k}{H_k + \lambda_{\text{reg}}}
\end{equation}
where $\eta$ is the learning rate and $\lambda_{\textbf{reg}} > 0$ is a regularization constant preventing excessive updates when curvature is small. Hence updated belief is obtained as:
\begin{equation}\label{eq:21}
\beta_{k+1} = \sigma \left ( \theta_{k+1} \right )
\end{equation}
Now, it is essential to test the proposed online-learning mechanism against the GNSS degradation model which is described in the next section.
\section{Urban Environment and GNSS Degradation Model}
This section defines a controlled degradation environment designed to evaluate the proposed second-order belief adaptation mechanism. The multipath and latency models are simplified, non-physical proxies introduced to induce non-stationary, correlated, and latency-coupled measurement degradation. The model is not intended to reproduce absolute GNSS statistics; rather, it captures the structural conditions under which conventional adaptive filters degrade. The objective is to assess estimator adaptation behavior under controlled non-stationarity, not to develop a high-fidelity GNSS propagation model, which is the focus in subsequent follow-up works. In order to model the effects of the GNSS degradation, we consider the quadrotor positions $\mathbf{p}_k = \left [ x_k, y_k \right ]^T$, while computing the minimum Euclidean distance to the nearest obstacle boundary given as,
\begin{equation} \label{eq:22}
	d_k = \min_i d \left ( \mathbf{p}_k, \mathcal{O}_i \right), 
\end{equation}
where $d \left ( \cdot \right )$ represents Euclidean distance. The scalar $d_k$ serves as a geometric proxy for the GNSS signal obstruction, obstacle reflection likelihood, and satellite visibility degradation. The satellite constellation or ray-tracing model is not considered here for analysis. Given the true GNSS observation $\mathbf{z}_k$,the multipath-induced bias $\nu_k$ is modelled as a zero-mean measurement noise as it evolves according to the first-order Gauss Markov process \cite{khanafseh2018gnss}:
\begin{equation}\label{eq:23}
	\mathbf{b}_{k+1} = \kappa \left (d_k \right) \mathbf{b}_{k} + \mathbf{w}_k, \quad \mathbf{w}_k \sim \mathcal{N} \left ( \mathbf{0}, \sigma_{b}^2 \left (d_k \right) \mathbf{I} \right),
\end{equation}
where  persistence coefficient $\kappa \left (d_k \right)$ and excitation variance $\sigma_{b}^2$ which depends on the obstacle proximity of the UAV $d_k$. The persistence coefficient is defined as,
\begin{equation}\label{eq:24}
	\kappa \left ( d_k \right) = \text{exp} \left ( - \frac{\Delta t}{\Gamma \left ( d_k \right)} \right )
\end{equation}
The correlation time coefficient is given as:
\begin{equation}\label{eq:25}
	\Gamma \left (d_k \right ) =
	\begin{cases}
		\Gamma_{\text{canyon}}, & d_k < d_{\text{near}}, \\
		\Gamma_{\text{open}},   & d_k \ge d_{\text{near}},
	\end{cases}
	\qquad
	\Gamma_{\text{canyon}} \gg \Gamma_{\text{open}}.
\end{equation}
The excitation variance is scaled as:
\begin{equation}\label{eq:26}
	\Gamma_b(d_k) =
	\begin{cases}
		\sigma_{\text{high}}, & d_k < d_{\text{near}}, \\
		\sigma_{\text{low}},  & d_k \ge d_{\text{near}}.
	\end{cases}
\end{equation}
The given formulation captures spatial dependence and temporal persistence aspect of the urban GNSS multipath signals. In addition to multipath bias, GNSS measurement noise increases in urban environments due to reduced visibility and degraded precision. This is modeled by inflating the noise covariance $\mathbf{R}_k$ by a factor $m_k$, with $m_k = r_{\text{urban}} > 1$ near buildings and $m_k = 1$ otherwise, which represents increased pseudorange uncertainty without satellite geometry model or signal-to-noise ratios. Urban degradation also introduces time-varying measurement latency due to receiver and tracking effects. The latency $L_k$ is modeled as a distance-dependent stochastic process with dynamics given by:
\begin{equation} \label{eq:27}
	L_{k+1} = \zeta L_k + \left ( 1- \zeta \right ) \bar{L} \left ( d_k \right ) + \Upsilon_k.
\end{equation}
Here, $\zeta \in \left (0,1 \right )$ controls the temporal correlation. The geometry-dependent mean latency is defined as, 
\[
\bar{L}(d_k) =
\begin{cases}
	L_{\text{canyon}}, & d_k < d_{\text{near}}, \\
	L_{\text{open}},   & d_k \ge d_{\text{near}},
\end{cases}
\qquad
L_{\text{canyon}} \gg L_{\text{open}}.
\]
The implementation of online autonomy framework under degraded GNSS measurements and time-varying latency is stated in Algorithm \ref{alg:belief_adaptive_ekf}.
\begin{algorithm}
	\caption{Second-order belief-adaptive EKF with latency-aware GNSS updates}
	\label{alg:belief_adaptive_ekf}
	\textbf{Input} IMU measurements $\mathbf{a}_k$, degraded GNSS measurements $\mathbf{\hat{z}}_k$, global waypoints $\mathcal{P}$, process and measurement models $\left (f, h \right )$, learning rates $(\eta, \lambda_{\text{reg}})$, curvature parameters $(\alpha_f, \alpha_s, \omega)$. \\
	\textbf{Output} Belief-adaptive state estimates $\hat{\mathbf{x}}_k$.
	\begin{algorithmic}[1]
		\State Initialize state estimate $\hat{\mathbf{x}}_0$, covariance $\mathbf{P}_0$
		\State Initialize belief parameter $\theta_0$, belief $\beta_0 = \sigma(\theta_0)$
		\State Initialize curvature accumulators $H_{\text{fast}}, H_{\text{slow}} > 0$
		
		\For{each control step $k$}
		\State $\hat{\mathbf{x}}_{k|k-1} \leftarrow f(\hat{\mathbf{x}}_{k-1|k-1}, \mathbf{a}_k)$
		\State $\mathbf{P}_{k|k-1} \leftarrow \mathbf{F}_k \mathbf{P}_{k-1} \mathbf{F}_k^\top + (1-\alpha^2) \sigma_b \left ( \beta_k \right)^2 \mathbf{I}$
		\If{delayed GNSS measurement $\mathbf{z}_{k-d}$ is available}
		\State Innovation $\mathbf{y} \leftarrow \mathbf{z}_{k-d} - \mathbf{H}\hat{\mathbf{x}}_{k-d}$
		\State $\mathbf{S} \leftarrow \mathbf{H}\mathbf{P}_{k-d}\mathbf{H}^\top + \mathbf{R}(\beta_{k-d})$
		\State $\mathbf{K} \leftarrow \mathbf{P}_{k-d}\mathbf{H}^\top \mathbf{S}^{-1}$
		\State $\hat{\mathbf{x}}_{k-d} \leftarrow \hat{\mathbf{x}}_{k-d} + \mathbf{K}\mathbf{y}$
		\State $\mathbf{P}_{k-d} \leftarrow (\mathbf{I}-\mathbf{K}\mathbf{H})\mathbf{P}_{k-d}$
		\State Re-propagate EKF forward to time $k$
		\State Compute NIS $\rho_k \leftarrow \mathbf{y}^\top \mathbf{S}^{-1} \mathbf{y}$
		\EndIf
		\State Compute the online learning loss as per (\ref{eq:16})
		\State Obtain the gradient for online belief update as per (\ref{eq:17}).
		\State Update $H_{\text{fast}}$ and $H_{\text{slow}}$ as per (\ref{eq:18}) to obtain $H_k$ in (\ref{eq:19}).
		\State Update the belief parameter as per (\ref{eq:20}).
		\State Obtain the updated belief parameter $\beta_{k+1}$ and update the measurement covariance matrix $\mathbf{Q}_b \left (\beta_{k+1} \right )$.
		\EndFor
	\end{algorithmic}
\end{algorithm}
\section{Results and Discussions}
\subsection{Numerical Implementation}
The proposed framework is implemented in Python environment while simulating the nonlinear quadrotor model as defined in (\ref{eq:7}) with full translational and rotational dynamics discretized at an IMU rate of 200 Hz ($\Delta t_{\text{imu}}$ = 0.005 s), while control and GNSS updates are applied at 20 Hz ($\Delta t_{\text{ctrl}}$ = 0.05 s). The quadrotor mass is set to 5 kg with inertia ${0.02, 0.02, 0.04}$ kg·m². IMU acceleration noise has standard deviation 0.08 $m/ s^2$, while nominal GNSS noise is 0.08 m and inflates to 0.1 m under multipath. GNSS multipath bias follows a first-order Gauss–Markov process with time constant $\Gamma_{\text{mp}}$ = 3 s and maximum bias standard deviation 1.2 m, scaled by obstacle proximity. Measurement latency is modeled as a geometry-dependent property with correlation time $\Gamma_{\text{delay}}$ = 2.5 s, noise standard deviation 0.03 s, and maximum delay capped at 1.2 s while inducing out-of-sequence updates.  Online belief adaptation uses a sigmoid-parameterized scalar belief updated via a dual-timescale second-order rule with learning rate $\eta = 0.15$ and regularization $\lambda_{\text{reg}} = 0.1$, driven by a hybrid loss combining normalized innovation squared ($\chi^2$ threshold at 95\%) and inter-step state variation. Performance is evaluated across multiple urban point-to-point missions and benchmarked against naive EKF, adaptive EKF, by modifying the measurement covariance matrix and first-order belief adaptation baselines.
\subsection{Simulation results}
\begin{figure}[thpb]
	\centering
	\includegraphics[scale=0.15]{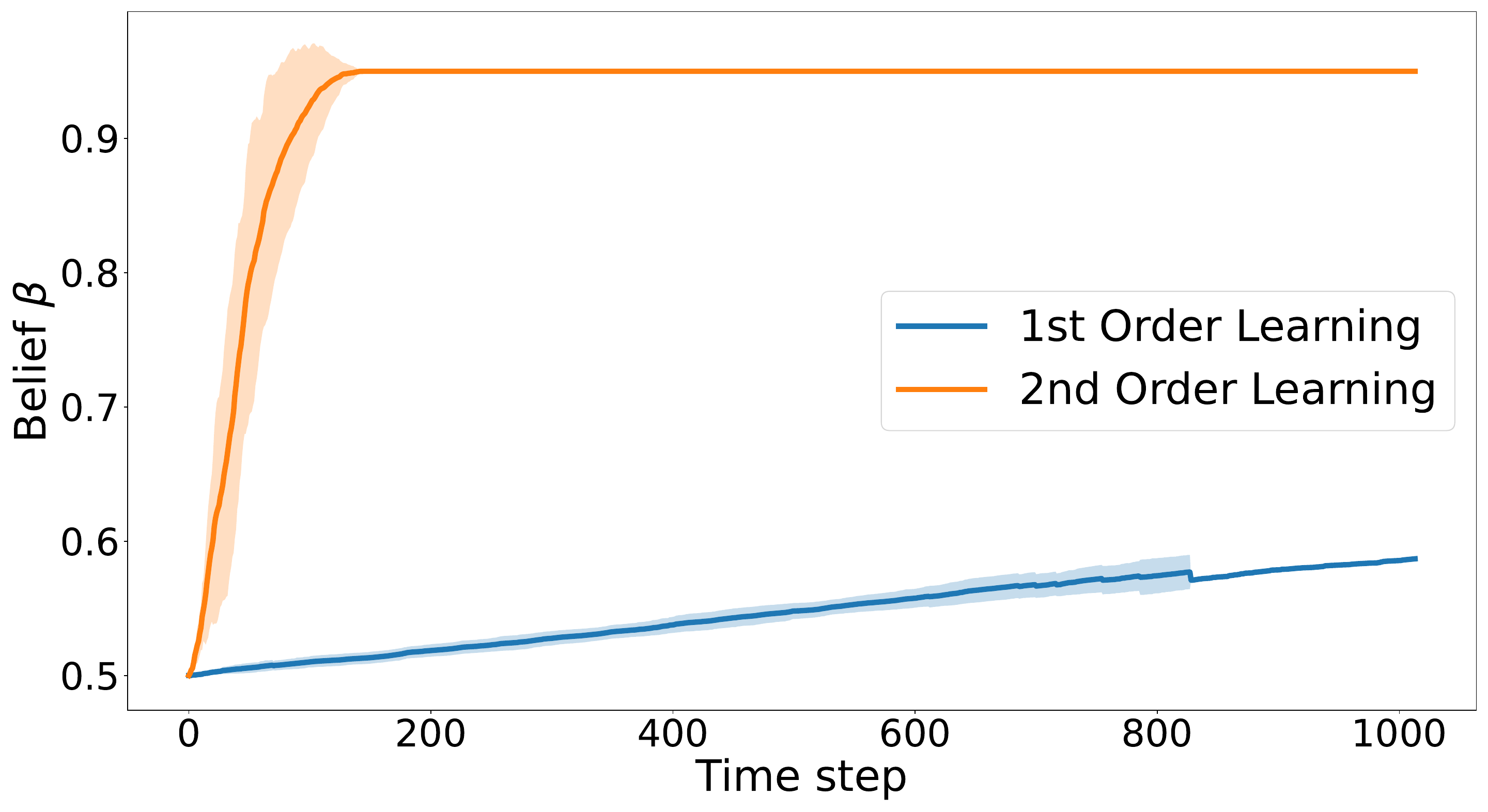}
	\caption{Comparison of belief adaptation over time for first-order and second-order online learning over 10 mission profiles.}
	\label{figure3_belief_convergence}
\end{figure}
\begin{figure}[thpb]
	\centering
	\includegraphics[scale=0.18]{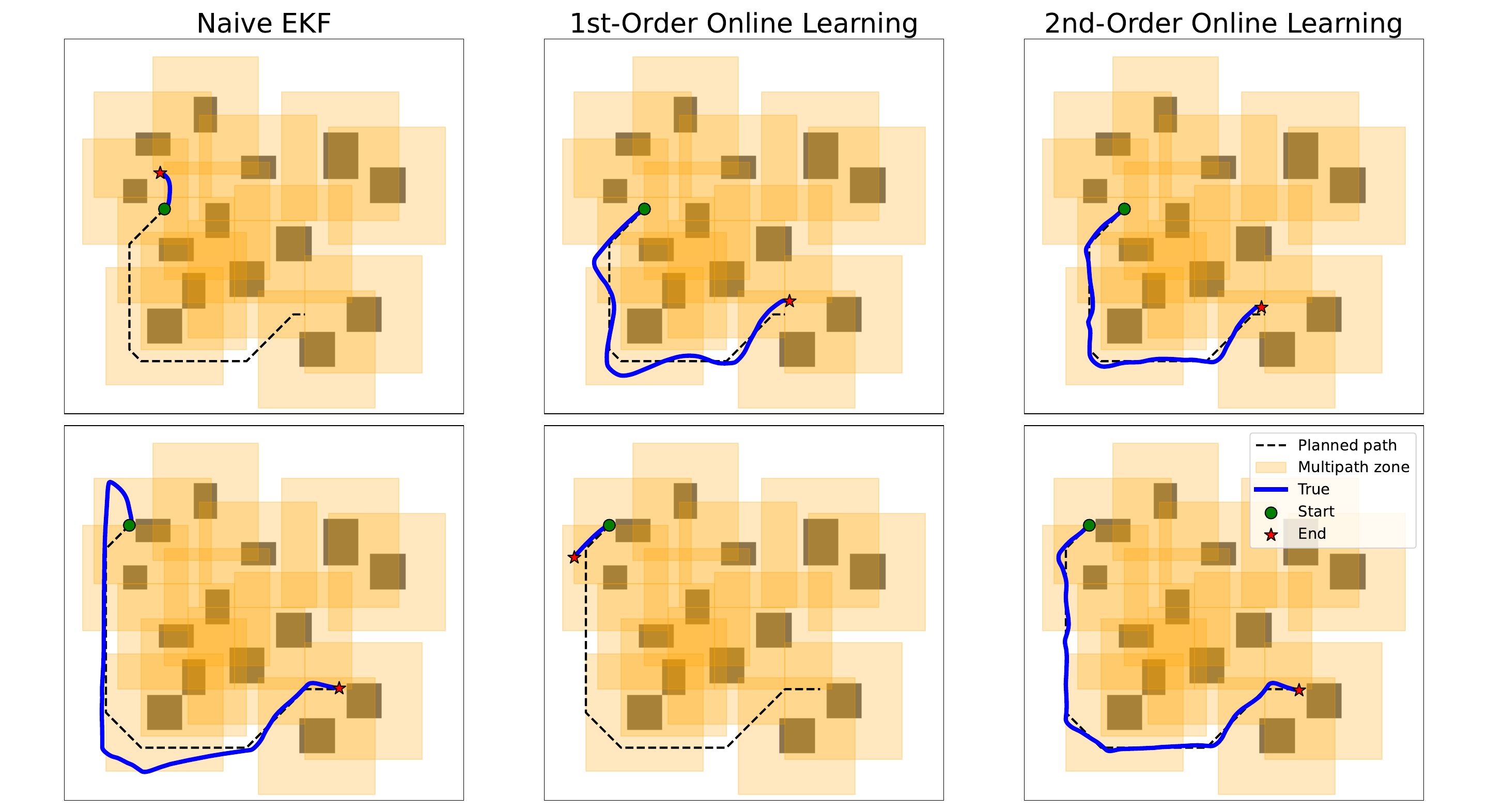}
	\caption{Comparison of quadrotor navigation trajectories under GNSS multipath and latency: naive EKF, first-order online learning, and second-order online learning. }
	\label{figure4_navigation_map}
\end{figure}
As demonstrated in Figure \ref{figure3_belief_convergence}, the belief state for second-order online learning achieves faster convergence to a stable trust level with reduced oscillation amplitude, while for first-order online learning a persistent drift and higher variability is observed. Trajectory comparisons in multipath-affected urban regions (Fig. \ref{figure4_navigation_map}) show that the naive EKF deviates from the planned path, particularly near waypoints located close to obstacles with strong multipath effects. The resulting unmitigated bias leads to degraded tracking of the reference trajectory from the global planner and increases the risk of unsafe navigation in obstacle-dense regions. First-order online learning reduces the overall deviation but exhibits delayed recovery and transient path-tracking errors, indicating limited responsiveness under degraded GNSS conditions. In the second mission, this leads to temporary violations of the planned path, as the controller relies primarily on local gradient information without accounting for global path constraints, as it moves out from the constrained airspace region. In contrast, second-order online learning maintains closer adherence to the planned trajectory and smoother motion through multipath-affected regions, consistently reaching the goal without large deviations. These results suggest that curvature-aware second-order belief adaptation provides  stable GNSS trust adjustment, leading to improved trajectory tracking and robustness under geometry-dependent multipath and latency effects.
\begin{figure}[thpb]
	\centering
	\includegraphics[scale=0.15]{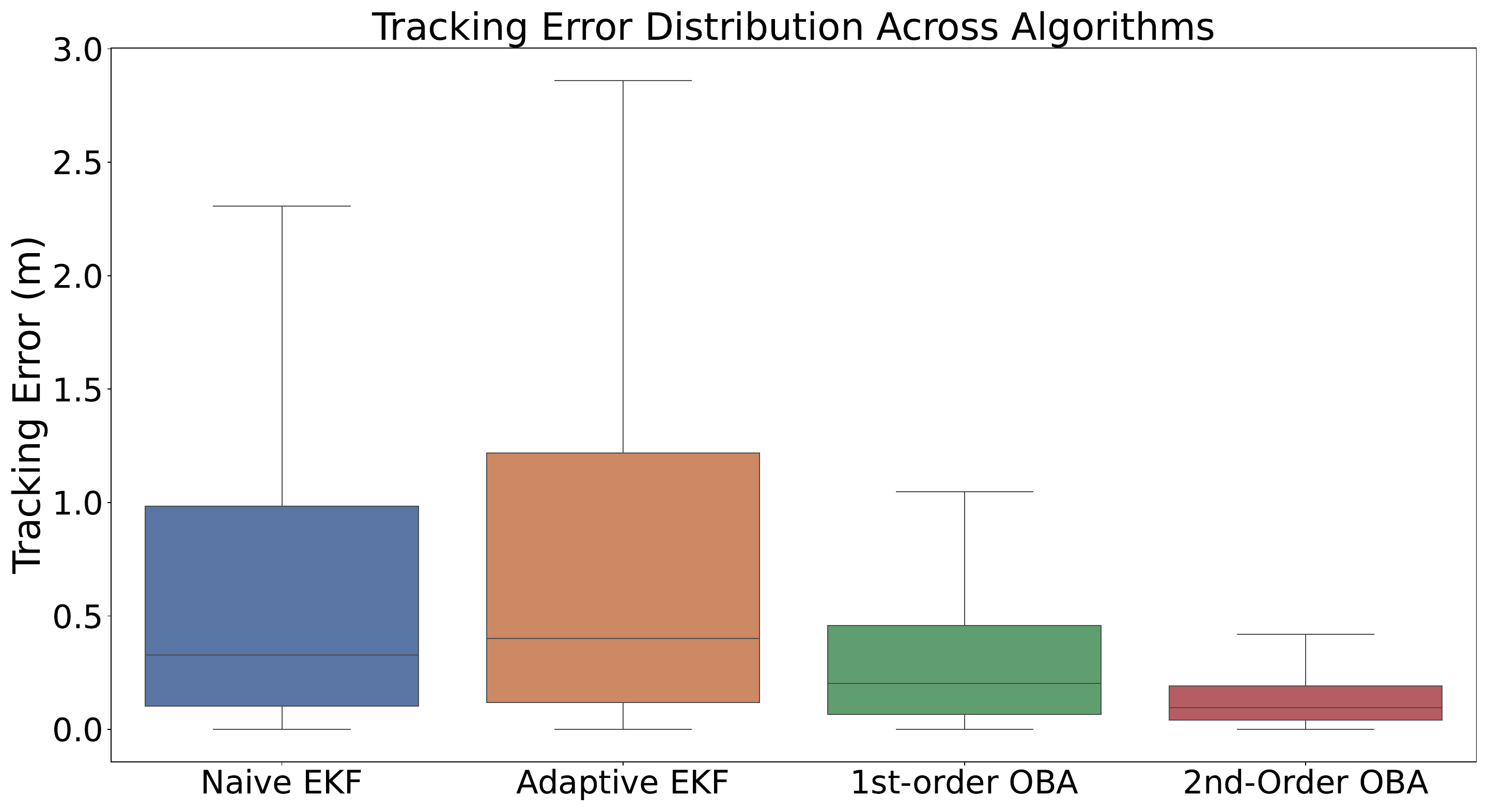}
	\caption{Tracking error distribution across classical and adaptive Kalman filter with respect to first and second order online learning methods over 10 mission profiles.}
	\label{figure5_tracking_error}
\end{figure}
\begin{figure}[thpb]
	\centering
	\includegraphics[scale=0.15]{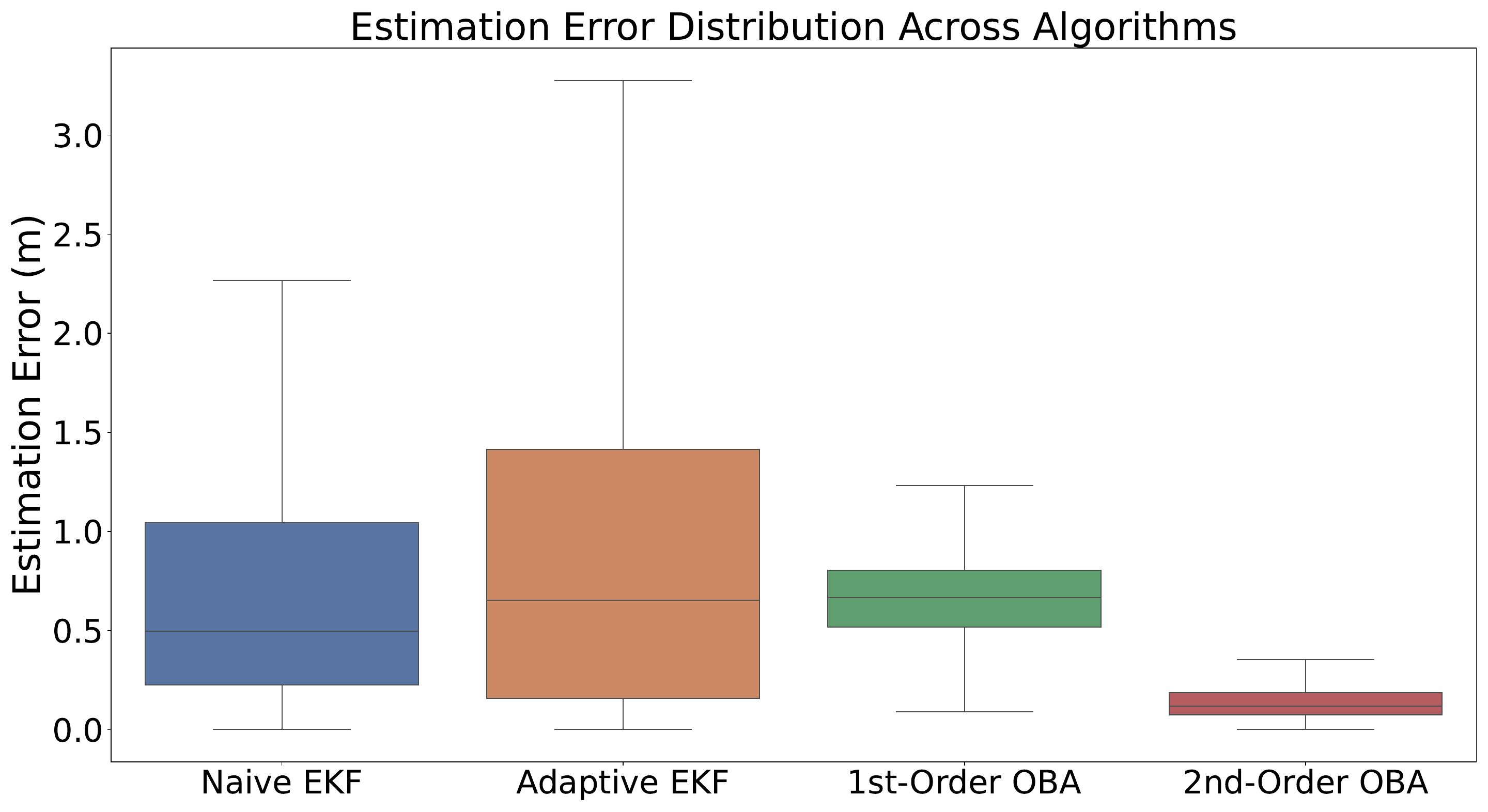}
	\caption{State estimation error distribution across classical and adaptive Kalman filter with respect to first and second order online learning methods over 10 mission profiles.}
	\label{figure6_estimation_error}
\end{figure}
We further analyze the results obtained across multiple mission profiles with different start and goal locations in the urban airspace, selected to ensure convergence of the global planner used to generate the reference waypoints. The improved belief stability of second-order online learning in Fig. \ref{figure3_belief_convergence} directly translates to better innovation stabilization leading to lower position estimation and trajectory tracking errors as shown  in Fig. \ref{figure5_tracking_error} and \ref{figure6_estimation_error}. The naive EKF exhibits the largest median errors and widest interquartile ranges, indicating high sensitivity to multipath bias and delayed measurements that violate its fixed-noise assumptions. The adaptive EKF reduces the median error but shows increased variance and long-tailed error distributions, reflecting reactive covariance tuning that struggles under rapidly changing GNSS conditions. Heavier tails observed in adaptive EKF reflect reactive covariance inflation rather than structured trust adaptation. First-order belief adaptation provides further improvement by explicitly modulating GNSS trust; however, its error distribution remains relatively dispersed, suggesting slower or inconsistent adaptation across missions. In contrast, the proposed second-order belief-adaptive method achieves the lowest median estimation with 80\% reduction in estimation errors and 30\% reduction in tracking errors across 10 mission profiles, demonstrating improved accuracy and consistency across all evaluated trajectories. These results demonstrate that curvature-aware adaptation reduces oscillatory trust updates, stabilizes innovation energy, and yields consistent covariance evolution across missions, even without modifying the underlying EKF structure.
\section{Conclusion}
This paper presented a belief-adaptive online autonomy framework for quadrotor navigation under GNSS degradation, integrating second-order online learning with a bias-augmented EKF and latency-aware measurement handling. By modeling GNSS reliability as a latent belief and adapting it through curvature-aware updates, the method enables stable sensor trust adjustment under geometry-dependent multipath and latency effects while navigating in urban environments. Simulation results show consistent improvements over naive and adaptive EKF, and first-order learning in terms of belief convergence, trajectory smoothness, and estimation and tracking accuracy. The framework preserves the structure of classical GNSS–IMU fusion and can be integrated into existing flight-control pipelines without offline training. Future work will focus on validation in real-world flight trials and extensions to multi-vehicle urban operations.
\printbibliography
\end{document}